\documentclass[runningheads]{llncs}

\usepackage{subcaption} 

\usepackage[T1]{fontenc}
\usepackage{graphicx}
\begin{document}
\title{Exploring the correlation between the type of music and the emotions evoked: A study using subjective questionnaires and EEG
}
%
%
\author{Jelizaveta Jankowska\inst{1} \and Bożena Kostek\inst{1} \and Fernando Alonso-Fernandez\inst{2} \and Prayag Tiwari\inst{2}}
\authorrunning{J. Jankowska et al.}
%
\institute{Audio Acoustics Lab - Faculty of Electronics, Telecommunications and Informatics, Gdańsk University of Technology (GUT), Poland \\ \email{jelkur23@student.hh.se, bozena.kostek@pg.edu.pl} \and School of Information Technology, Halmstad University, Sweden \\ \email{feralo@hh.se, prayag.tiwari@hh.se}}
\maketitle              
\begin{abstract}

The subject of this work is to check how different types of music affect human emotions. While listening to music, a subjective survey and brain activity measurements were carried out using an EEG helmet. The aim is to demonstrate the impact of different music genres on emotions. The research involved a diverse group of participants of different gender and musical preferences. This had the effect of capturing a wide range of emotional responses to music. After the experiment, a relationship analysis of the respondents’ questionnaires with EEG signals was performed. The analysis revealed connections between emotions and observed brain activity.

\keywords{Music and Emotion  \and EEG-Based Emotion Recognition \and \and Brain-Computer Interface (BCI).}
\end{abstract}

\section{Introduction}

Music, as an art form based on sounds, has been an integral part of human existence, transforming and evolving over time. 
Neurological studies revealed that music has the ability to release dopamine, the happiness hormone, and can evoke diverse emotions, inﬂuencing the perception of time \cite{ref1}.
Advancements in brain activity research allowed for a scientiﬁc approach to understanding how music affects the human psyche and physiology \cite{ref4}.

Different types of music can affect human emotions, as observed through EEG recordings. Studies have shown that music across genres, including Indian classical music (ICM) and melodies played with different instruments, can evoke a variety of emotions \cite{ref5,ref6}. EEG experiments have been conducted to assess the neural responses of both musicians and non-musicians to different emotional stimuli in music, revealing that musicians tend to exhibit higher complexity and coherence in their brain responses compared to non-musicians \cite{ref7}. 
Furthermore, EEG-based emotion recognition has shown promising results in accurately identifying and classifying different emotions using music audio signals.
Additionally, the impact of music stimuli on the brain and its physiological function system has been studied, providing insights into the role of music in clinical treatment and neurological interventions \cite{ref8}. 

But despite advancements in technology and insights from previous studies, signiﬁcant gaps remain in the understanding of how music affects individuals. While AI and brainwave technology offer potential paths for recognising people's emotions, there has been limited exploration into how various music styles relate to emotional states and physiological responses \cite{ref9,ref10,ref11}.
Accordingly, this research aims to ﬁnd answers to questions regarding the impact of music on human emotions. It involves investigating the correlation between different types of music and the emotions they evoke, employing questionnaires and EEG signal recordings. 
The objective is to establish a relationship between participants' subjective responses in questionnaires regarding their emotional experiences and the objective EEG signal data during music listening, determining whether there is a visible connection between subjective emotional perceptions and physiological responses measured through EEG signals. 

\section{Background}

\subsection{Study of Emotions}
\label{sect:emotions}

\begin{figure}[t]
\centering
\includegraphics[width=0.48\textwidth]{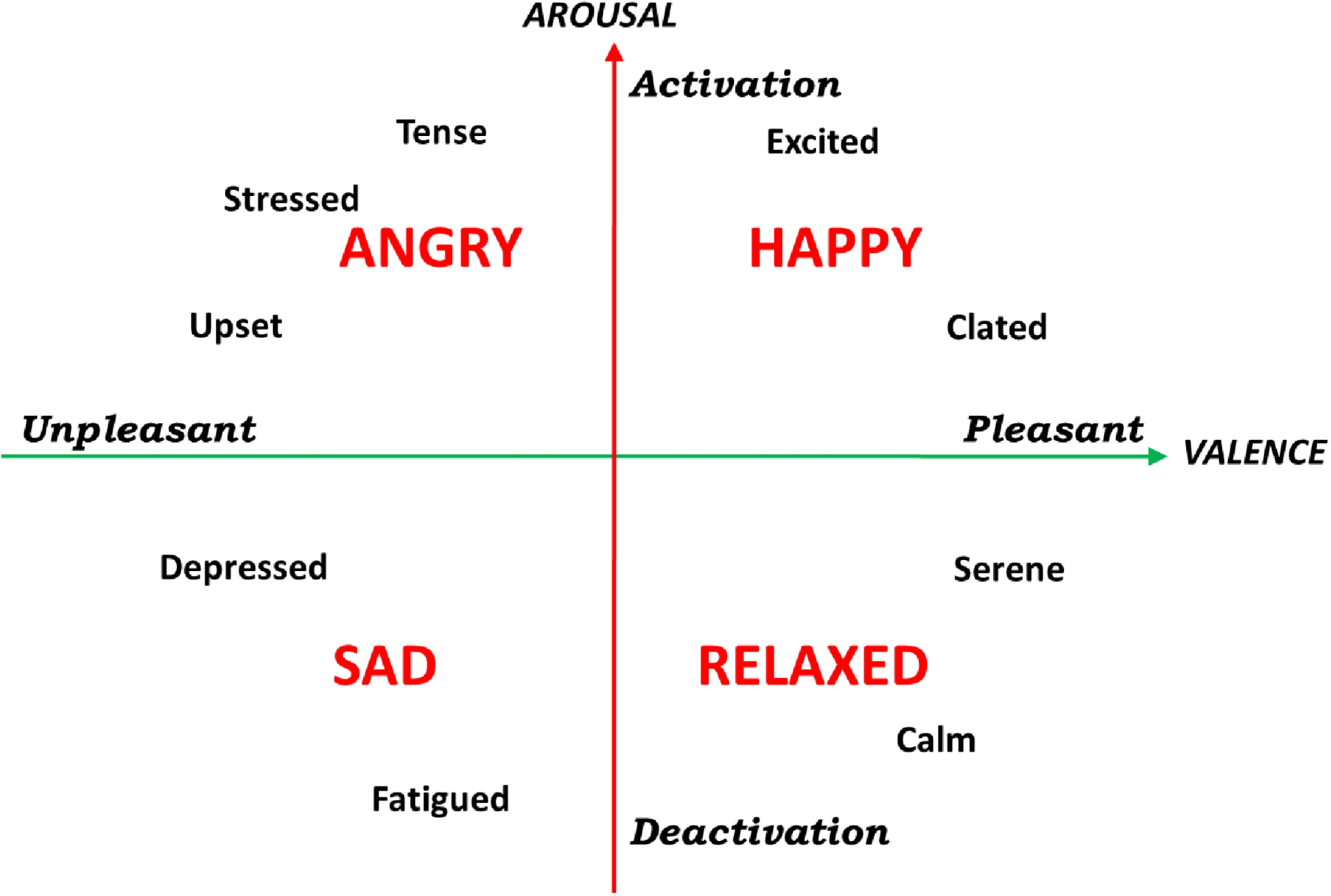}
\includegraphics[width=0.45\textwidth]{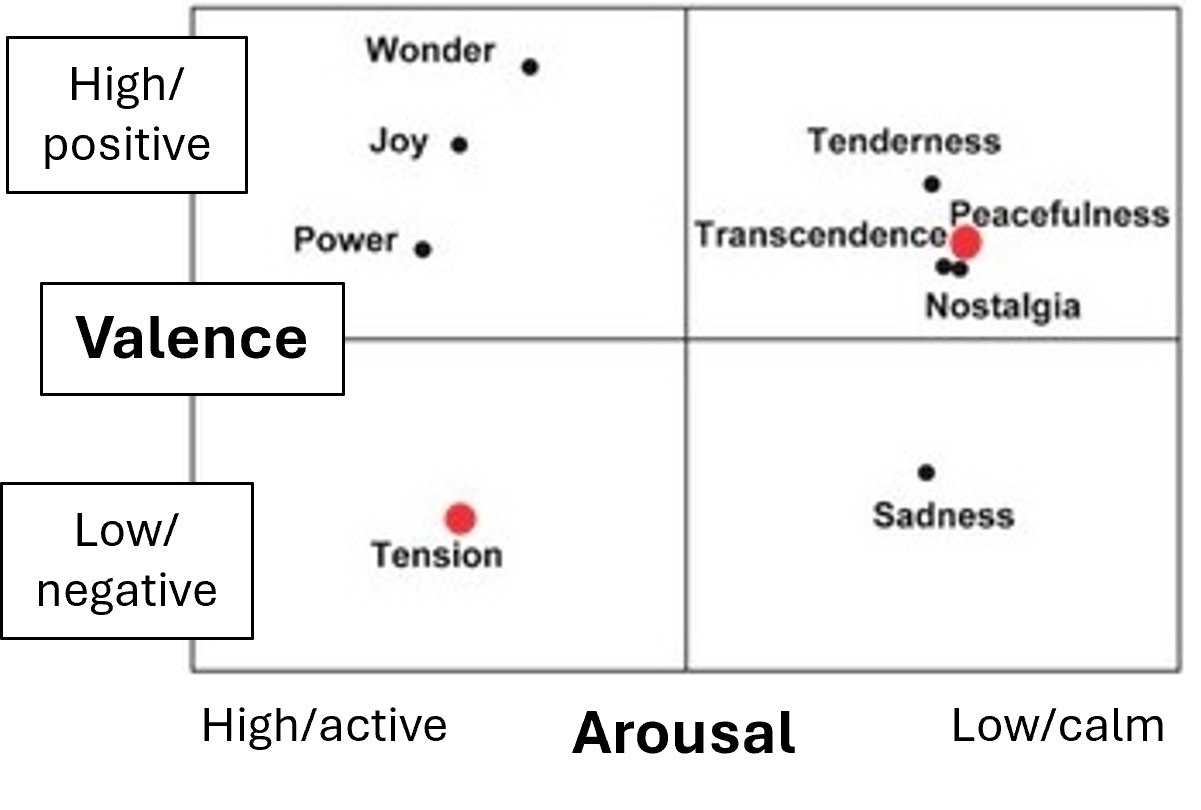}
\caption{Left: Russel’s model of emotions based on valence-arousal scale \cite{ref34}. Right: Geneva Emotional Music Scale (GEMS) \cite{ref57} and mapping to the V-A scale.} \label{fig:fig36}
\end{figure}

Emotional states encompass various phenomena such as feelings, affects, motivations, moods, passions, and sentiments, which, though theoretically distinct, are closely interrelated in practice \cite{ref12}.
Emotions are defined as complex processes involving arousal, subjective experience, physiological changes, expressions, and behavioural tendencies \cite{ref16}.
Two dominant theories explain emotional emergence: Ekman’s theory of six universal emotions (happiness, sadness, anger, fear, disgust, and surprise) \cite{ref18}, identifiable through facial and body expressions and from which all other emotions derive; and Barrett’s constructivist approach \cite{ref20}, which views emotions as culturally shaped and individually variable experiences.
A widely adopted framework in affective science is Russell’s circumplex model \cite{ref34} (Figure~\ref{fig:fig36}, left), which maps emotions along valence (pleasure vs. displeasure) and arousal (activation or alertness vs. deactivation) axes. Additional variables like intensity (the extent to which the process inﬂuences behaviour, thought processes, etc.) and content (meaning of the stimulus and predisposition to speciﬁc behaviours such as fear leading to escape, or anger leading to aggressiveness) can be used to characterize emotional states further \cite{ref17}. 

Emotion recognition methods can be based on physiological or non-physiological signals \cite{ref26}. 
Non-physiological approaches include analysis of written or spoken language, speech, and body behaviour, often using self-report tools like questionnaires or structured interviews. 
Physiological methods, on the other hand, rely on objective signals such as EEG, ECG, facial muscle activity, eye movement, and skin conductance. In principle, these are not susceptible to subjective factors or interpretation, offering more reliable and accurate assessments of emotional states.
To elicit and study emotions, researchers use diverse stimuli: static images, auditory inputs (music, speech), videos, and immersive Virtual Reality (VR) environments, each one enabling observation of emotional responses through both self-reports and physiological monitoring.

\subsection{Electroencephalography (EEG)}
\label{sect:eeg_soa}

Electroencephalography (EEG) is a non-invasive technique for measuring spontaneous brain electrical activity, with amplitudes typically between 1–100 $\mu$V and frequencies from 0 to 100 Hz. 
Unlike MRI or PET, EEG offers millisecond-level temporal resolution. 
EEG is primarily used for diagnosing and monitoring neurological conditions.
In scientiﬁc research, it is utilised in neuroscience, cognitive psychology, neurolinguistics, and psychophysiological studies. 
It is also increasingly applied to emotion research \cite{ref26,ref31}, due to the relation between activated frequency bands and brain states:

\begin{itemize}
    \item Delta (0.5–3 Hz): deep sleep; linked to neurological disorders and depression.

    \item Theta (4–7 Hz): light sleep, meditation, cognitive processing, daydreaming.
    
    \item Alpha (8–12 Hz): relaxed wakefulness, calmness, positive mood; decreases with attention or mental effort.

    \item Beta (13–28 Hz): active thinking, learning, and emotional arousal, both pleasant or unpleasant (e.g. anger or agitation).

    \item Gamma (>30 Hz, up to 100 Hz, with the relevant information below 50 Hz): high-level cognition, positive meditative states, gratitude, love, and joy.
\end{itemize}

EEG offers portability and high resolution, operating quietly, and it is relatively affordable compared with other neuroimaging techniques.
Despite advantages, it has limitations too.
It is sensitive to movement and muscle artefacts such as blinking, requiring signal filtering. 
In addition, the preparation process can be lengthy and uncomfortable. Conductive gel may be needed, which can be inconvenient, with dry EEG methods offering less accuracy.

\section{Methodology and Results}

The methodology includes several key components, which are addressed in the following subsections:
$\textbf{i)}$ the creation of a music database aimed at evoking various emotional responses;
$\textbf{ii)}$ the analysis of musical parameters of the selected music samples;
$\textbf{iii)}$ the development of a structured subjective survey to capture participants' emotional reactions;
$\textbf{iv)}$ the execution of EEG experiments using EEG device to gather brain activity data;
$\textbf{v)}$ the processing of this EEG data to extract meaningful insights, exploring the relationship between subjective emotional experiences and objective EEG-based measures.

\begin{table}[t]
\centering
\caption{Left: total counts per emotion/music genre of Emotify. Centre/right: the calmest tracks of classical music/the most tense tracks of electronic music.}
\label{tab:emotify-analysis}
\resizebox{0.95\textwidth}{!}{%
\begin{tabular}{ccccccccccc}
\multicolumn{1}{l}{} & \multicolumn{1}{l}{} & \multicolumn{1}{l}{} & \multicolumn{1}{l}{} & \multicolumn{1}{l}{} & \multicolumn{1}{l}{} & \multicolumn{1}{l}{} & \multicolumn{1}{l}{} & \multicolumn{1}{l}{} & \multicolumn{1}{l}{} & \multicolumn{1}{l}{} \\ \cline{1-5} \cline{7-8} \cline{10-11} 
\multicolumn{1}{|c|}{\textbf{\begin{tabular}[c]{@{}c@{}}Emotion\\ (GEMS)\end{tabular}}} & \multicolumn{1}{c|}{\textbf{Classical}} & \multicolumn{1}{c|}{\textbf{Rock}} & \multicolumn{1}{c|}{\textbf{Electronic}} & \multicolumn{1}{c|}{\textbf{Pop}} & \multicolumn{1}{c|}{} & \multicolumn{1}{c|}{\textbf{\begin{tabular}[c]{@{}c@{}}Classical\\ track  ID\end{tabular}}} & \multicolumn{1}{c|}{\textbf{\begin{tabular}[c]{@{}c@{}}Average\\ calmness\end{tabular}}} & \multicolumn{1}{c|}{} & \multicolumn{1}{c|}{\textbf{\begin{tabular}[c]{@{}c@{}}Electronic\\ track ID\end{tabular}}} & \multicolumn{1}{c|}{\textbf{\begin{tabular}[c]{@{}c@{}}Average\\ tension\end{tabular}}} \\ \cline{1-5} \cline{7-8} \cline{10-11} 
\multicolumn{1}{|c|}{Wonder} & \multicolumn{1}{c|}{423} & \multicolumn{1}{c|}{253} & \multicolumn{1}{c|}{224} & \multicolumn{1}{c|}{219} & \multicolumn{1}{c|}{} & \multicolumn{1}{c|}{93} & \multicolumn{1}{c|}{88\%} & \multicolumn{1}{c|}{} & \multicolumn{1}{c|}{260} & \multicolumn{1}{c|}{82\%} \\ \cline{1-5} \cline{7-8} \cline{10-11} 
\multicolumn{1}{|c|}{Transcendence} & \multicolumn{1}{c|}{694} & \multicolumn{1}{c|}{288} & \multicolumn{1}{c|}{425} & \multicolumn{1}{c|}{265} & \multicolumn{1}{c|}{} & \multicolumn{1}{c|}{99} & \multicolumn{1}{c|}{88\%} & \multicolumn{1}{c|}{} & \multicolumn{1}{c|}{262} & \multicolumn{1}{c|}{82\%} \\ \cline{1-5} \cline{7-8} \cline{10-11} 
\multicolumn{1}{|c|}{Tenderness} & \multicolumn{1}{c|}{558} & \multicolumn{1}{c|}{385} & \multicolumn{1}{c|}{132} & \multicolumn{1}{c|}{476} & \multicolumn{1}{c|}{} & \multicolumn{1}{c|}{82} & \multicolumn{1}{c|}{81\%} & \multicolumn{1}{c|}{} & \multicolumn{1}{c|}{229} & \multicolumn{1}{c|}{73\%} \\ \cline{1-5} \cline{7-8} \cline{10-11} 
\multicolumn{1}{|c|}{Nostalgia} & \multicolumn{1}{c|}{676} & \multicolumn{1}{c|}{\textbf{630}} & \multicolumn{1}{c|}{238} & \multicolumn{1}{c|}{\textbf{627}} & \multicolumn{1}{c|}{} & \multicolumn{1}{c|}{4} & \multicolumn{1}{c|}{76\%} & \multicolumn{1}{c|}{} & \multicolumn{1}{c|}{245} & \multicolumn{1}{c|}{73\%} \\ \cline{1-5} \cline{7-8} \cline{10-11} 
\multicolumn{1}{|c|}{Calmness} & \multicolumn{1}{c|}{\textbf{919}} & \multicolumn{1}{c|}{518} & \multicolumn{1}{c|}{509} & \multicolumn{1}{c|}{615} & \multicolumn{1}{c|}{} & \multicolumn{1}{c|}{2} & \multicolumn{1}{c|}{74\%} & \multicolumn{1}{c|}{} & \multicolumn{1}{c|}{254} & \multicolumn{1}{c|}{73\%} \\ \cline{1-5} \cline{7-8} \cline{10-11} 
\multicolumn{1}{|c|}{Power} & \multicolumn{1}{c|}{388} & \multicolumn{1}{c|}{458} & \multicolumn{1}{c|}{459} & \multicolumn{1}{c|}{226} &  &  &  &  &  &  \\ \cline{1-5}
\multicolumn{1}{|c|}{Joy} & \multicolumn{1}{c|}{748} & \multicolumn{1}{c|}{484} & \multicolumn{1}{c|}{516} & \multicolumn{1}{c|}{397} &  &  &  &  &  &  \\ \cline{1-5}
\multicolumn{1}{|c|}{Tension} & \multicolumn{1}{c|}{464} & \multicolumn{1}{c|}{400} & \multicolumn{1}{c|}{\textbf{693}} & \multicolumn{1}{c|}{333} &  &  &  &  &  &  \\ \cline{1-5}
\multicolumn{1}{|c|}{Sadness} & \multicolumn{1}{c|}{464} & \multicolumn{1}{c|}{437} & \multicolumn{1}{c|}{109} & \multicolumn{1}{c|}{434} &  &  &  &  &  &  \\ \cline{1-5}
\end{tabular}%
}
\end{table}

\begin{figure}[t]
\centering
\includegraphics[width=0.9\textwidth]{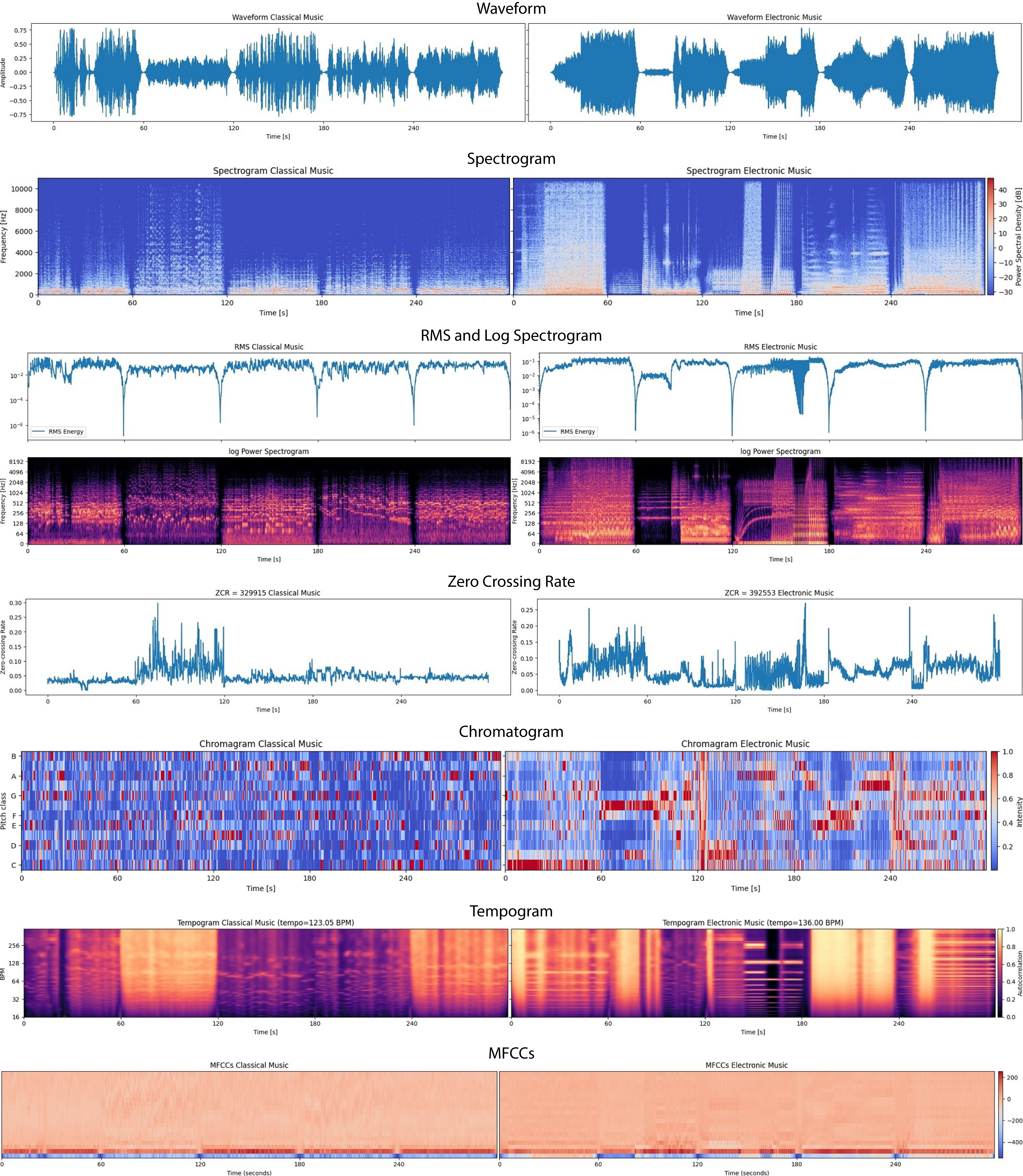} 
\caption{Musical parameters of classical (left column) and electronic music (right).} \label{fig:musical_parameters}
\end{figure}

\subsection{Database Creation}
\label{sect:database}

For this study, a customised dataset of music fragments was created based on open-source data collections.
We use a valence-arousal scale classification method \cite{ref34}. 
For such purpose, we employed the Emotify dataset \cite{ref57} as base, which comprises 400 music fragments, each lasting one minute, categorised into four genres: rock, classical, pop, and electronic. Each genre contains 100 music fragments, and the emotional annotations were collected using the GEMS scale \cite{ref55}. 
The Geneva Emotional Music Scale (GEMS) is a model designed specially to measure musical emotions, identifying nine fundamental emotions experienced by humans while listening to music, which can be mapped onto the valence-arousal (V-A) scale (Figure~\ref{fig:fig36}, right). 
When listening, participants of Emotify were asked to select up to 3 emotions of the GEMS scale that they felt were evoked by the music fragments. 
To analyze the dataset, we calculated the total counts per emotion for every music genre (Table~\ref{tab:emotify-analysis}, left).
The results reveal that classical music predominantly elicits positive emotions, with high counts for emotions such as calmness (peacefulness), indicating its calming and uplifting effects. 
Conversely, electronic music most commonly induces tension, with the highest count for this emotion among all genres. 
Rock and pop music both strongly evoke nostalgia, as reﬂected in their high scores for this emotion.
Tension and calmness are opposing emotions on the V-A scale (red dots in Figure~\ref{fig:fig36}, right).
The prevalence of calmness in classical music and tension in electronic music suggests that these genres have signiﬁcantly different emotional impacts.

Based on the above, we decided to compare classical and electronic music. 
Since the dataset comprises one-minute fragments, we created longer recordings by concatenating several fragments with similar emotional qualities. 
To do so, we identified the ﬁve most calming classical and the most tension-inducing electronic tracks (Table~\ref{tab:emotify-analysis}, center, right). 
The selected tracks were arranged sequentially, with fade-in and fade-out between them to avoid abrupt changes, resulting in two 5-minute files, one classical and one electronic. 

\subsection{Musical Parameters}
\label{sect:parameters}

Music is composed of various elements that interact to create a wide variety of sounds, including melody, harmony, rhythm, texture, timbre, dynamics, form, and tempo \cite{ref45}.
We analyzed various audio parameters of the created 5-minute tracks to understand the impact of musical elements on listener emotions.
Figure~\ref{fig:musical_parameters} presents their waveform, spectrogram, RMS, zero crossing rate (ZCR), chromatogram, tempogram and Mel-frequency cepstral coefﬁcients (MFCCs).

The waveforms reveal characteristic patterns. Classical music displays repetitive, cyclical structures with prominent peaks corresponding to the ﬁrst beat of each measure. In contrast, electronic music has more irregular/unpredictable high peaks. 
Notably, the second classical track (seconds 60-120), featuring a solo piano, exhibited lower amplitude compared to other tracks.

The spectrogram is a heat map representation of frequencies' intensity over time. In classical music, lower frequencies dominate, with high frequencies having lower power spectral density. Electronic music shows sharper and more unpredictable changes, as observed in the waveform previously, and reflected here by higher frequency energy at individual times. 

Root-mean-square (RMS) measures the total magnitude of the signal, often interpreted as the loudness or energy. Classical music has lower RMS ($\sim$10$^{-2}$) compared to electronic music ($\sim$10$^{-1}$), consistent with their respective loudness and energy levels seen in the spectrogram. The RMS graphs, aligned with logarithmic spectrograms, highlight these differences further. 

Zero crossing rate (ZCR) measures the change rate of the signal from positive to negative or vice-versa. Higher ZCR values are common in percussive sounds. ZCR for classical music is relatively stable (0.05-0.1), except for the second piano track (0.2-0.3). Electronic music exhibits more variability (0.02-0.25), conﬁrming its percussive and dynamic nature.

The chromatogram shows the dominance of pitches (e.g., C, D, E, etc.) in the audio. Classical music exhibits harmonic blends of tones, while electronic music displays more inconsistent tonal shifts. 

The tempogram visualises the tempo, or speed, of the audio piece, measured in beats per minute (bpm). Upbeat genres tend to have higher tempos. A relatively similar average tempo for both genres is observed (123 vs. 136 bpm). However, classical music shows smoother patterns across time, while electronic music has distinct, sharp lines reflective of sudden changes in tempo. 

Finally, Mel-frequency cepstral coefficients (MFCCs) represent the short-term power spectrum of a sound on a Mel scale, useful for distinguishing different sounds, especially speech. MFCC show negligible differences between classical and electronic music, as both recordings lack signiﬁcant vocal components.

\begin{table}[t]
\centering
\caption{STOMP analysis for every participant. Bold indicates the highest count per ID.}
\label{tab:STOMP}
\begin{tabular}{ccccc}
 &  &  &  &  \\ \hline
\multicolumn{1}{|c|}{\textbf{ID}} & \multicolumn{1}{c|}{\textbf{\begin{tabular}[c]{@{}c@{}}Reﬂective \& \\ Complex\end{tabular}}} & \multicolumn{1}{c|}{\textbf{\begin{tabular}[c]{@{}c@{}}Intense \& \\ Rebellious\end{tabular}}} & \multicolumn{1}{c|}{\textbf{\begin{tabular}[c]{@{}c@{}}Upbeat \& \\ Conventional\end{tabular}}} & \multicolumn{1}{c|}{\textbf{\begin{tabular}[c]{@{}c@{}}Energetic \& \\ Rhythmic\end{tabular}}} \\ \hline
\multicolumn{1}{|c|}{1} & \multicolumn{1}{c|}{\textbf{22}} & \multicolumn{1}{c|}{15} & \multicolumn{1}{c|}{21} & \multicolumn{1}{c|}{10} \\ \hline
\multicolumn{1}{|c|}{2} & \multicolumn{1}{c|}{\textbf{24}} & \multicolumn{1}{c|}{20} & \multicolumn{1}{c|}{23} & \multicolumn{1}{c|}{17} \\ \hline
\multicolumn{1}{|c|}{3} & \multicolumn{1}{c|}{17} & \multicolumn{1}{c|}{15} & \multicolumn{1}{c|}{\textbf{18}} & \multicolumn{1}{c|}{16} \\ \hline
\multicolumn{1}{|c|}{4} & \multicolumn{1}{c|}{\textbf{23}} & \multicolumn{1}{c|}{19} & \multicolumn{1}{c|}{22} & \multicolumn{1}{c|}{10} \\ \hline
\multicolumn{1}{|c|}{5} & \multicolumn{1}{c|}{\textbf{23}} & \multicolumn{1}{c|}{21} & \multicolumn{1}{c|}{22} & \multicolumn{1}{c|}{19} \\ \hline
\multicolumn{1}{|c|}{6} & \multicolumn{1}{c|}{13} & \multicolumn{1}{c|}{9} & \multicolumn{1}{c|}{\textbf{20}} & \multicolumn{1}{c|}{18} \\ \hline
\multicolumn{1}{|c|}{7} & \multicolumn{1}{c|}{13} & \multicolumn{1}{c|}{14} & \multicolumn{1}{c|}{15} & \multicolumn{1}{c|}{\textbf{17}} \\ \hline
\multicolumn{1}{|c|}{8} & \multicolumn{1}{c|}{18} & \multicolumn{1}{c|}{\textbf{20}} & \multicolumn{1}{c|}{19} & \multicolumn{1}{c|}{13} \\ \hline
\multicolumn{1}{|c|}{9} & \multicolumn{1}{c|}{17} & \multicolumn{1}{c|}{\textbf{18}} & \multicolumn{1}{c|}{17} & \multicolumn{1}{c|}{15} \\ \hline
\multicolumn{1}{|c|}{10} & \multicolumn{1}{c|}{18} & \multicolumn{1}{c|}{12} & \multicolumn{1}{c|}{\textbf{20}} & \multicolumn{1}{c|}{17} \\ \hline
\end{tabular}
\end{table}

\begin{figure}[t]
\centering
\includegraphics[width=0.98\textwidth]{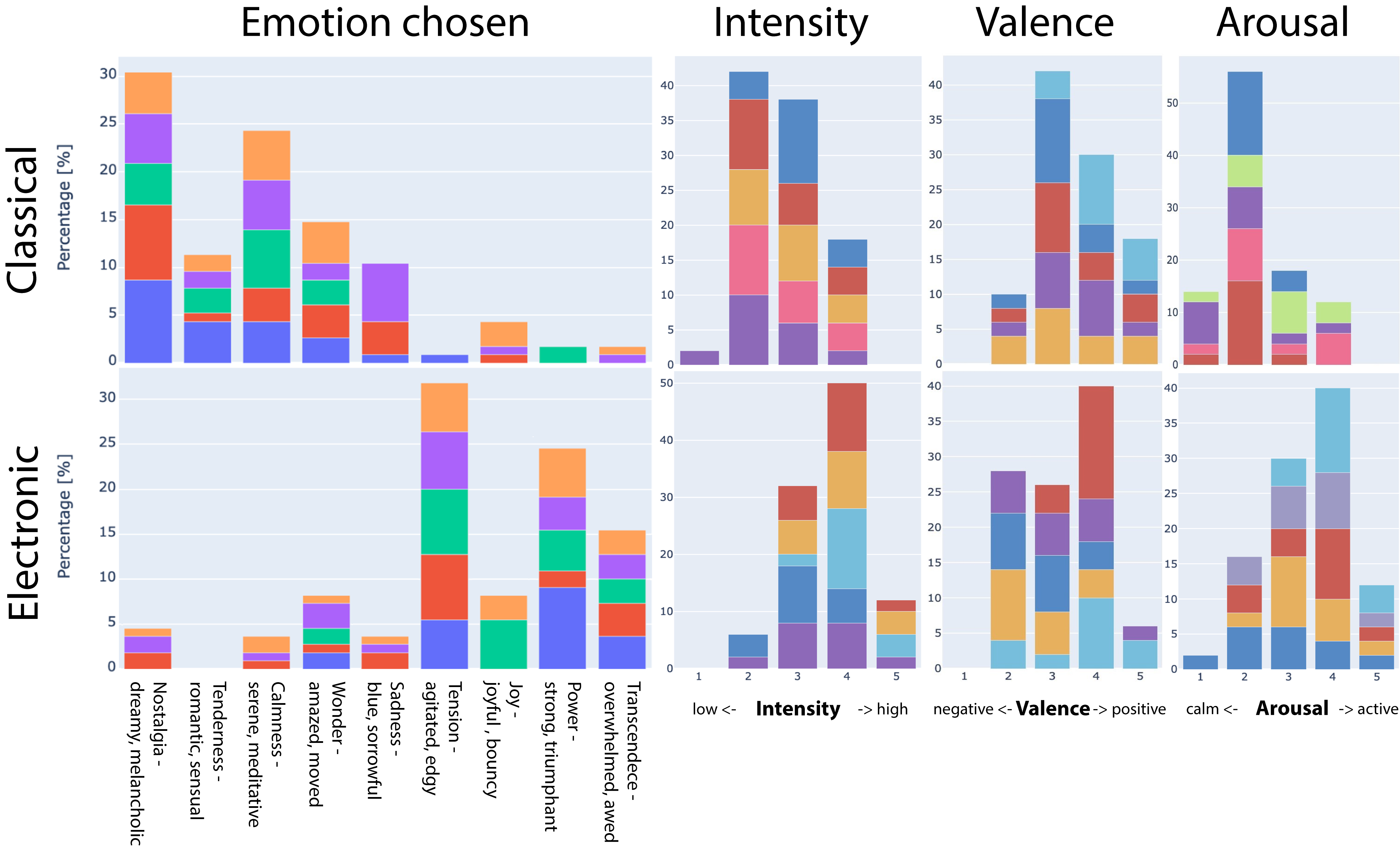}
\caption{Questionnaire results.} \label{fig:fig41_48}
\end{figure}

\subsection{Survey Development and Results}
\label{sect:survey}

A structured subjective survey (available at \cite{ref59}) 
was formulated to elicit participants' emotional responses to music stimuli. 
We created a Google Form with embedded YouTube video links containing the audio tracks.
An introductory section collects participant information (gender, age), music preferences and music background. Two sections with four questions each followed to rate emotional responses to classical and electronic music, respectively, after listening to the tracks. 
We engaged 7 male and 3 female participants aged 20-29 from (**hidden for double blind**).
Users were asked to select up to 3 GEMS emotions that felt while listening to each track, how intense they were (1-5), the valence level (1-5, from negative to positive), and arousal level (1-5, from calm to active).

Nine out of ten participants reported enjoying music, with one being ambivalent. 
Only one participant had musical training. No formal musical training was preferred to avoid professional bias in both EEG and survey results.
Musical preferences were assessed with the STOMP (Short Test of Musical Preferences) questionnaire \cite{ref60}, which involves rating 14 music genres on a scale from 1 to 7. 
Then, they are grouped into 4 types (Table~\ref{tab:STOMP}, with the highest value per participant in bold). 
The results show a wide variety of participants, with 4 belonging to the Reflective \& Complex category, 2 to Intense \& Rebellious, 3 to Upbeat \& Conventional, and 1 to Energetic \& Rhythmic. 
Interestingly, participants in the Reﬂective \& Complex category also showed high scores in the Upbeat \& Conventional (1-2 points difference only).

Figure~\ref{fig:fig41_48} shows the results of the emotional responses answers.
Regarding the selected emotion (first column), nostalgia was the most dominant response to classical tracks, reported by 30\% of respondents. Tranquility (25\%) and wonder (15\%) were the next most frequently reported.
Nostalgia and tranquility are close in the GEMS V-A mapping (Figure~\ref{fig:fig36}, right), which aligns with our criteria for selecting tracks from Emotify, since the classical tracks chosen were those scoring the highest in tranquility (calmness).
Regarding intensity (second column), emotions elicited by classical music were not very intense, with 42\% selecting '2' and 38\% selecting '3'. This suggests that classical music induces milder emotional responses. For valence (third column), classical music was perceived as neutral to positive, with 42\% choosing '3' and 30\% choosing '4'. No participant rated classical music as negative ('1'). In terms of arousal (fourth column), classical music was rated as more calming, with 55\% selecting '2'. Scores of '1', '3', '4', and '5' did not exceed 20\%, and no participant selected '5'.
In summary, classical music was perceived by participants as inducing calmness and tranquility, with mild emotional intensity, a more positive than neutral valence, and decidedly low arousal, indicating a calming effect.

Electronic music, on the other hand, elicited different emotional responses.
Tension was the predominant emotion (31\%), in line with our track selection criteria, followed by power (25\%) and transcendence (15\%).
The contradiction between tension and power (they have opposite valence, Figure~\ref{fig:fig36}, right) suggests that electronic music can be perceived both positively and negatively, depending on the listener.
For valence, opinions were more varied, reflecting these oppositions. While 40\% rated it as '4' (positive), 28\% rated it as '2' (negative), and 25\% rated it as '3' (neutral). In terms of arousal, electronic music was perceived as more stimulating, with 40\% selecting '4' and 30\% selecting '3'. This contrasts with classical music, where '2' was dominant. Regarding the intensity of emotions, electronic music induced more intense emotions, with 50\% selecting '4' and 32\% selecting '3'. Despite listening to the recordings for the ﬁrst time, electronic music elicited more intense emotional responses.


\begin{figure}[t]
\centering
\includegraphics[width=0.48\textwidth]{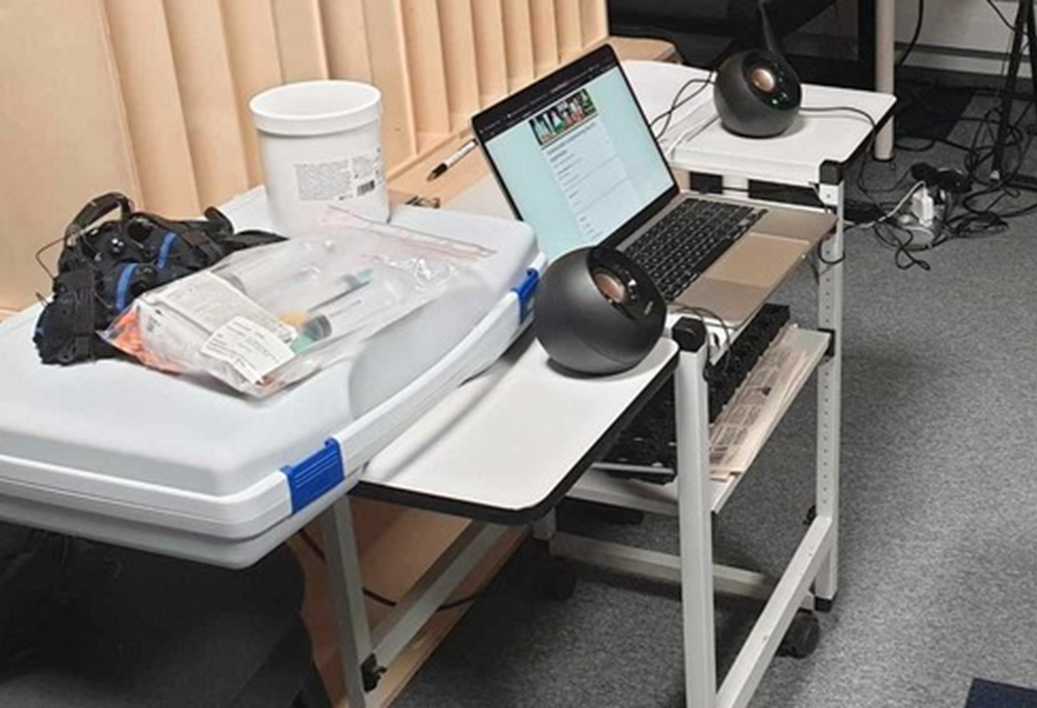}
\includegraphics[width=0.48\textwidth]{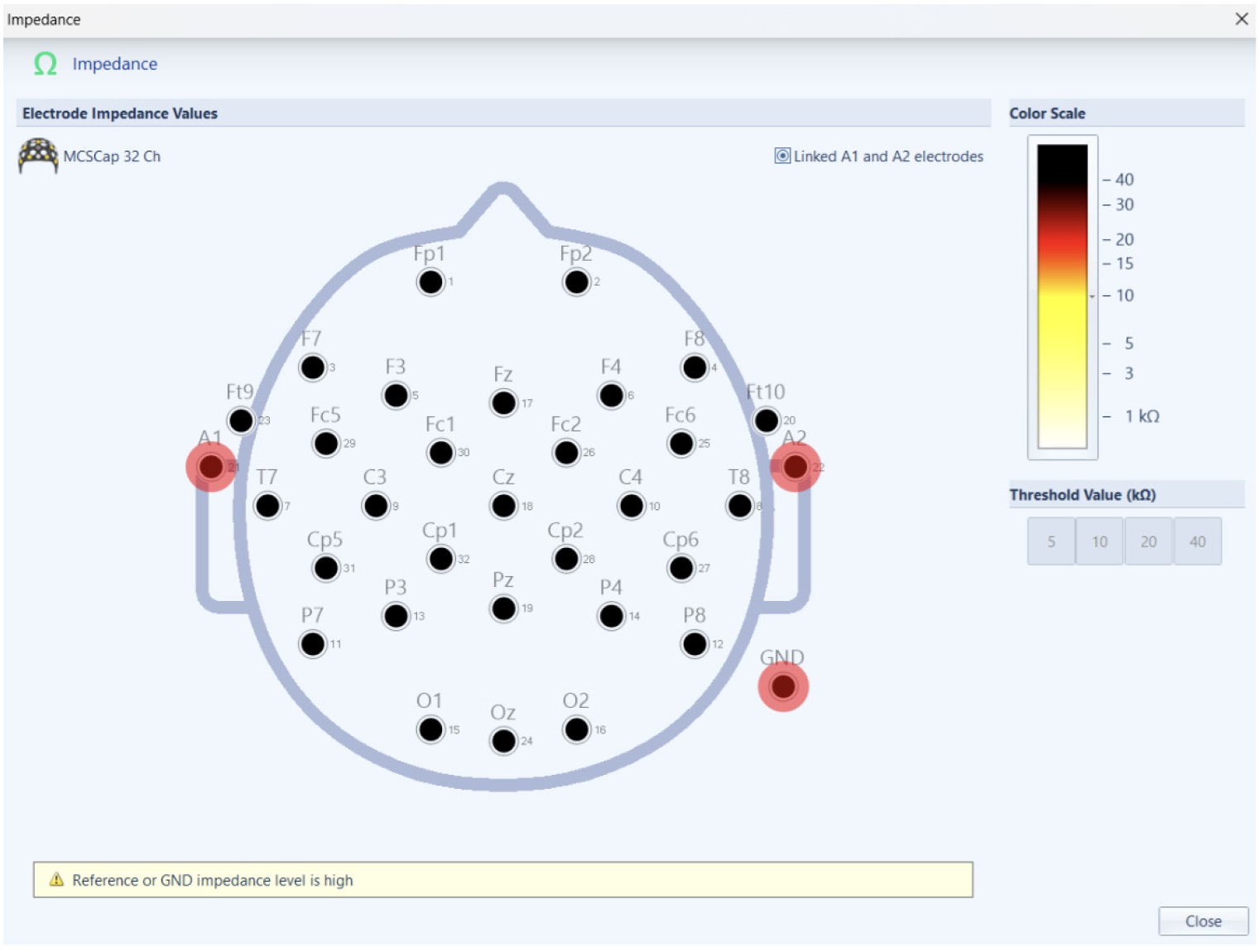}
\caption{EEG experimentation. Left: equipment. Right: Electrodes conductivity (yellow-white colour preferable).} \label{fig:fig14bc}
\end{figure}

\begin{figure}[t]
\centering
\includegraphics[width=0.98\textwidth]{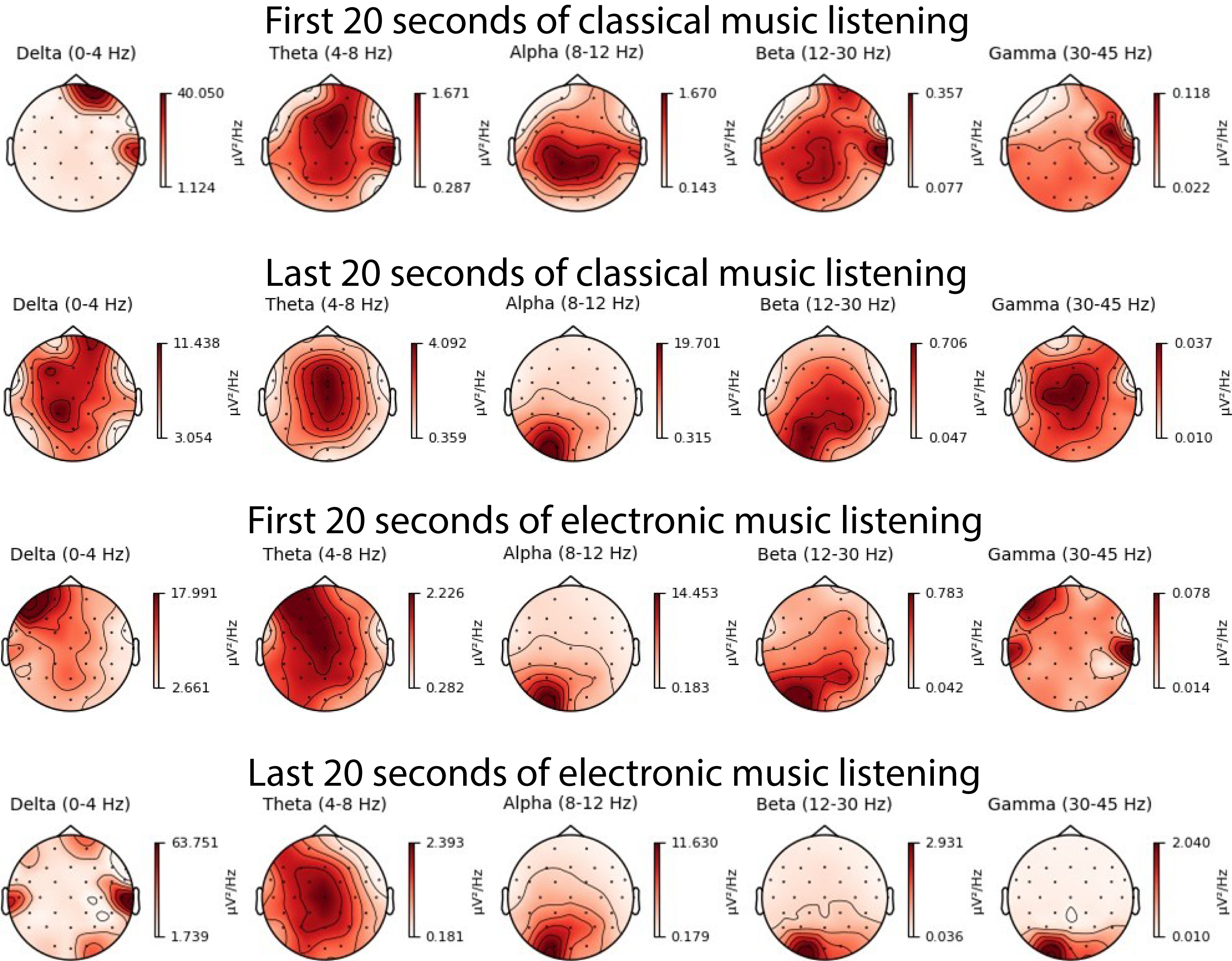}
\caption{Topomap of EEG data at the beginning and end of classical and electronic music listening.} \label{fig:fig57}
\end{figure}

\subsection{EEG Experiments and Analysis}
\label{sect:eeg_experiments}

For the EEG studies, we employed the MITSAR medical measurement device \cite{ref61}, with 32 electrodes. To ensure no disturbances, the experiment was conducted in a sound-isolated room at **hidden for double blind** (Figure~\ref{fig:fig14bc}, left). Comfortable seating and high-quality speakers were set up to ensure comfortable music listening. 
The EEG cap was fitted by applying gel to the electrodes to achieve a low resistance level below 5kOhm (Figure~\ref{fig:fig14bc}, right). 
Participants first completed the general section of the questionnaire.
EEG recording started a few seconds before playing classical music to capture the participant's baseline state, followed by 5 minutes of classical music playback, and a few seconds of resting state. 
Then, participants filled out the section to rate emotional responses to classical music.
Afterwards, the recording restarted to play the electronic music section, concluding with another short period of rest.

To process EEG data, we applied filtering to remove frequencies below 0.1Hz and above 45Hz. We then applied Independent Component Analysis (ICA) to remove artefacts due to eye movements, blinking, etc. 
Figure~\ref{fig:fig57} shows the average topographic maps of the users with the EEG activity across the scalp for the first and last 20 seconds of classical and electronic music listening. 
Recall that theta and alpha waves are associated with relaxation and meditation. In contrast, beta and gamma are associated with intense thinking and action, and delta is associated with tensional states or disorders.
Examining the topomaps, we see that before listening to classical music (first row), there was tension in the frontal delta region (forehead and eyes), which dissipated at the end (second row, observe the different scales). 
Then, theta and alpha power increased (relaxation), beta slightly increased, and gamma decreased. 
Before listening to electronic music (third row), frontal dental tension increased again, and it was further amplified to very high levels in the lateral regions after the session, maybe indicating jaw clenching.
Theta and alpha (relaxation) remained medium-low and even decreased during the electronic session. In contrast, beta and gamma (action) increased significantly.

\section{Conclusions}

This study aims to explore how different music genres impact human emotions. During the experiments, subjective surveys and EEG measurements were conducted using music fragments identified to be calm-inducing (classical music) and tension-inducing (electronic music).

The analysis revealed correlations between music genres and emotional responses. The research involves participants of various backgrounds, providing a range of emotional reactions to music. 
The survey results indicated that classical music is associated with calm and tranquil emotions, lower emotional intensity, positive valence, and low arousal. In contrast, electronic music elicits a wider range of emotions, including tension and power, higher arousal, and more intense emotional responses. This diversity in emotional reactions highlights the subjective nature of music perception.

Regarding EEG analysis, it showed that classical music promotes relaxation, increasing theta and alpha wave activity while decreasing frontal tension. In contrast, electronic music increases delta tension and beta and gamma activity, suggesting a more stimulating and arousing effect. The combination of subjective survey responses and objective EEG data provided a comprehensive understanding of the emotional and neural impact of different music genres on individuals.

Future work will involve the study of machine learning algorithms to characterize emotions automatically by EEG \cite{RecenaMenezes2017} and assess if they match with the self-reported answers. 
We are also working on including more participants across a wider age range to better generalize our findings, as well as providing more granularity in music genres by incorporating tracks with other styles.


 


\section*{Acknowledgements} 

This work has been carried out by J. Jankowska in the context of her Master degree in Biomedical/Medical Engineering at Gdańsk University of Technology. 
The work was partly done while J. Jankowska was an exchange student at Halmstad University (Computer Science and Engineering program).
F. A.-F. thanks the Swedish Research Council (VR) for funding his research.

%
%
%
\bibliographystyle{splncs04}
%

\bibliography{refs_eeg_music}

\end{document}